\documentclass{Interspeech2024}
\usepackage{pythonhighlight}
\usepackage{multirow}
\usepackage{amsmath}

\interspeechcameraready

\title{Lightweight Transducer Based on Frame-Level Criterion}

\name[affiliation={1,2}]{Genshun}{Wan}
\name[affiliation={2}]{Mengzhi}{Wang}
\name[affiliation={2}]{Tingzhi}{Mao}
\name[affiliation={1,*}]{Hang}{Chen}
\name[affiliation={1}]{Zhongfu}{Ye}

\address{
  $^1$\thanks{*Corresponding author}University of Science and Technology of China, China\\
  $^2$iFLYTEK Research, China}
\email{\{gswan,wmz1995\}@mail.ustc.edu.cn, tzmao@iflytek.com, \{hangchen,yezf\}@ustc.edu.cn}


\begin{document}

\maketitle

\begin{abstract}
    
    The transducer model trained based on sequence-level criterion requires a lot of memory due to the generation of the large probability matrix. We proposed a lightweight transducer model based on frame-level criterion, which uses the results of the CTC forced alignment algorithm to determine the label for each frame. Then the encoder output can be combined with the decoder output at the corresponding time, rather than adding each element output by the encoder to each element output by the decoder as in the transducer. This significantly reduces memory and computation requirements. To address the problem of imbalanced classification caused by excessive blanks in the label, we decouple the blank and non-blank probabilities and truncate the gradient of the blank classifier to the main network. Experiments on the AISHELL-1 demonstrate that this enables the lightweight transducer to achieve similar results to transducer. Additionally, we use richer information to predict the probability of blank, achieving superior results to transducer.
\end{abstract}

\section{Introduction}

Nowadays, end-to-end speech recognition models are gradually becoming mainstream \cite{watanabe2018espnet, Zhang2022WeNet2M, gao22b_interspeech, Radford2022RobustSR}. Compared with traditional hybrid models, end-to-end models have simpler training and higher recognition accuracy. End-to-end models mainly include CTC \cite{Graves2006ConnectionistTC}, LAS \cite{Chan2015ListenAA}, and transducer \cite{Graves2012SequenceTW}. CTC is a model with conditional independence assumptions, while LAS and transducer are non-conditional independence assumptions. Models with conditional independence assumptions do not consider the historical output in prediction and do not have a good modeling of semantic dependencies, so their accuracy is lower than that of the models of non-conditional independence assumptions. The LAS model is based on attention mechanism, although the attention mechanism can solve various sequence-to-sequence tasks, it is not suitable for speech recognition tasks because it does not consider the monotony of speech and may lead to abnormal phenomena \cite{Chiu2019ACO} such as repetitive decoding. Although there are variations like mocha attention \cite{Chiu2017MonotonicCA} that improve the monotonicity of attention, the recognition accuracy also decreases. Transducer is a frame-by-frame decoding model like CTC without decoding anomalies. Unlike the CTC model, the transducer solves the problem of conditional independence using the prediction network and the joint network, but at the cost of producing a huge probability matrix, with a memory consumption that is an order of magnitude larger than that of CTC and LAS models.

Some efforts restrict the set of possible paths within a certain region of an alignment when calculating transducer loss to reduce memory usage. The difference is how to obtain the alignment. It can be CTC \cite{Wang2022AcceleratingRT}, a small RNN-T \cite{Kuang2022PrunedRF}, or continuous integrate-and-fire (CIF) \cite{An2023BATBA}. Their commonality is that they are still based on sequence-level criterion, which is not easy to implement. For instance, although both our work and \cite{Wang2022AcceleratingRT} utilize CTC guidance, they partition the T-by-U lattice into strips and employ sequence-level criterion, whereas we employ frame-level criterion. This eliminates the need for lattice segmentation, simplifying implementation and boosting training efficiency without strip-based pruning. Furthermore, while they discard frames with CTC blank probabilities above a threshold, we use the CTC forced alignment algorithm to determine blank frames, which is more accurate.

Another approach abandons transducer loss and combines label-level encoder representations extracted from CIF with prediction network outputs \cite{Deng2023LabelSynchronousNT,Zhang2023SayGT}. This approach is essentially a variant of CIF and cannot be called the transducer, as they discard the blank symbol in the transducer and rely on the accumulation of CIF quantities for alignment during decoding. Since the CIF-based model is a soft frame-synchronous model that needs to locate the acoustic boundary during processing, this makes it perform slightly inferior on the datasets with blurred acoustic boundary between labels as discussed in \cite{Dong2020ACO}.

In this paper, we proposed a lightweight transducer model that preserves the blank symbol and does not rely on CIF for alignment during decoding. Therefore, our model is more robust to audio with blurred acoustic boundary. We use the results of the CTC forced alignment algorithm as the frame-level label. Since the label for each frame is determined, we only need to combine the encoder output with the decoder output at the corresponding time. Specifically, the first frame of the encoder output is combined with the initial decoder state. If the first frame corresponds to a blank symbol, the decoder state remains unchanged. Otherwise, the decoder state advances one step. Then the decoder state combines with the second frame. The subsequent encoder frames are processed in the same manner. As a result, memory usage reduces from \(O(N*T*U*V)\) to \(O(N*T*V)\), where \(N\) is the batch size, \(T\) is the output length of the encoder, \(U\) is the output length of the prediction network, and \(V\) is the vocabulary size.  Besides, we can train the lightweight transducer using frame-level cross-entropy loss. This simplifies the engineering implementation and results in a lower training cost compared to sequence-level criterion. In experiments, we found that the naive implementation performed poorly, mainly due to the imbalance of multi-classification caused by excessive blanks in frame-level labels. By decoupling blank and non-blank probabilities and truncating the gradient of the blank classifier to the main network, we successfully trained the lightweight transducer with similar accuracy to the transducer. In addition, we analyzed when the model should output a blank and provide more information to the blank classifier, achieving superior results to transducer. We validated our approach on the AISHELL-1 \cite{Bu2017AISHELL1AO} dataset. The implementation is available on GitHub\footnote{https://github.com/wangmengzhi/Lightweight-Transducer}.


\section{Transducer}

\begin{figure}[t]
  \centering
  \includegraphics[width=\linewidth]{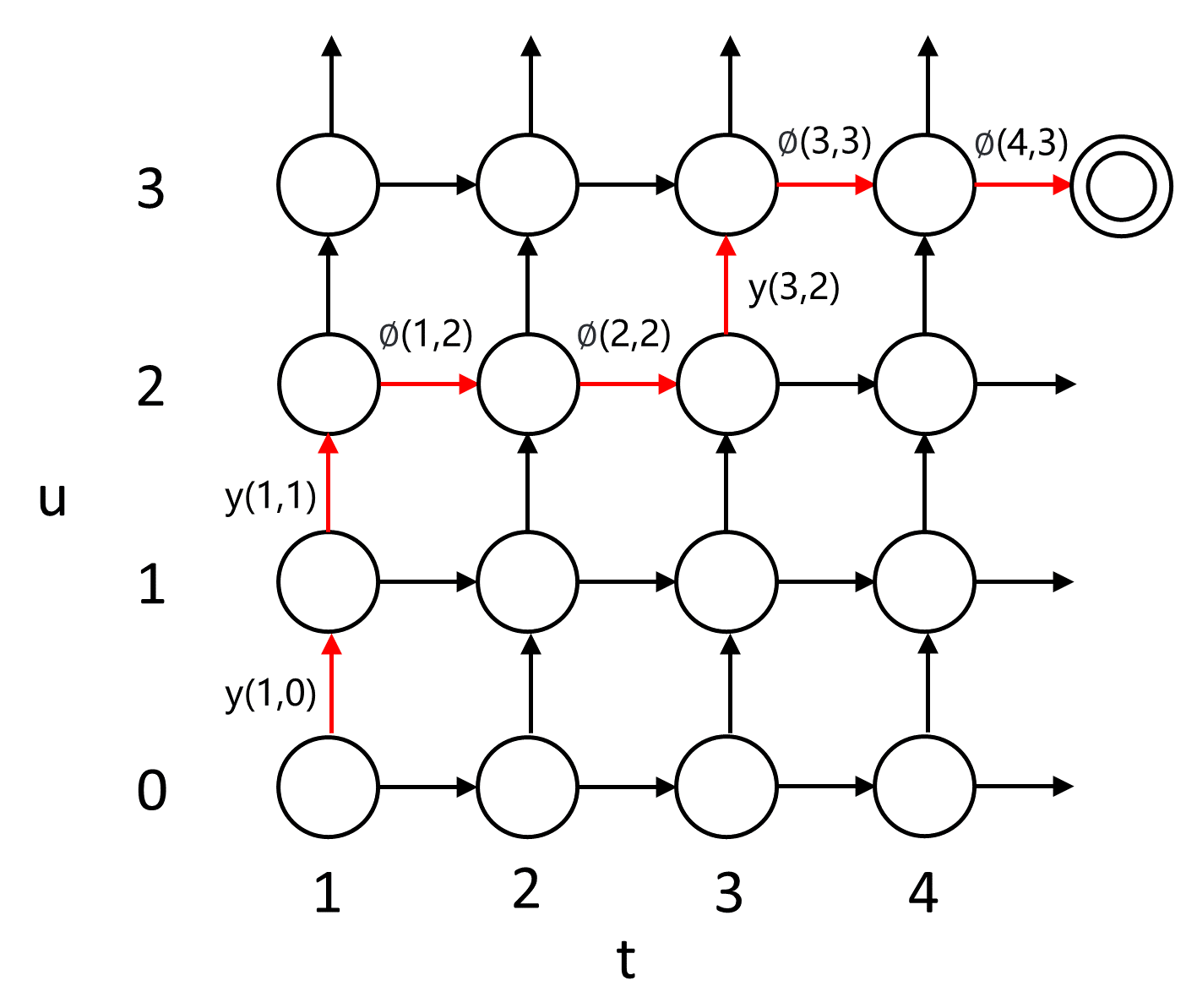}
  \caption{Output probability lattice defined by \(Pr(k|t,u)\). The node at \((t, u)\) represents the probability of having output the first \(u\) elements of the output sequence by point \(t\) in the transcription sequence. The horizontal arrow leaving node \((t, u)\) represents the probability \(\emptyset(t,u)\) of outputting nothing at \((t, u)\); the vertical arrow represents the probability \(y(t,u)\) of outputting the element \(u + 1\) of \(y\).}
  \label{fig:lattice}
\end{figure}

The transducer contains an encoder, a decoder (prediction network), and a joint network. The encoder encodes speech frames \(x = (x_1,...,x_T  ) \) into high-level acoustic features, similar to the acoustic model. The decoder encodes historical text \(y = (y_1,...,y_U  )\) into high-level language features, similar to the language model. In the joint network, each acoustic feature and each language feature are added together to generate an output probability distribution \(Pr(k|t,u)\), where \(k\) denotes the k-th category, \(t\) denotes the t-th encoder output and \(u\) denotes the u-th decoder output. \(Pr(k|t,u)\) is used to determine the transition probabilities in the lattice shown in Fig. \ref{fig:lattice}. Each path from bottom left to top right represents an alignment between \(x\) and \(y\). The sum of the log probabilities of each edge on the path is the path probability, and the sum of the probabilities of all paths is the total probability \(Pr(y|x)\) of the output sequence given the input sequence. The transducer loss uses forward backward algorithm to calculate \(Pr(y|x)\). Let the forward variable \(\alpha(t,u)\) be the log-probability outputting \(y_{0..u}\) after seen \(x_{0..t}\). It can be computed recursively using \eqref{alpha}
\begin{equation}\label{alpha} 
  \alpha(t,u)=\alpha(t-1,u)\emptyset(t-1,u)+\alpha(t,u-1)y(t,u-1)
\end{equation}
The output shape of the joint network is \((N, T, U, V)\), which is much larger than \((N, T, V)\) or \((N, U, V)\) required by CTC and LAS. Therefore, transducer training requires a large amount of memory. To avoid out of memory, the batch size must be reduced, thus leading to slower training.

\section{Lightweight transducer}

\begin{figure}[t]

  \centering
  \includegraphics[width=\linewidth]{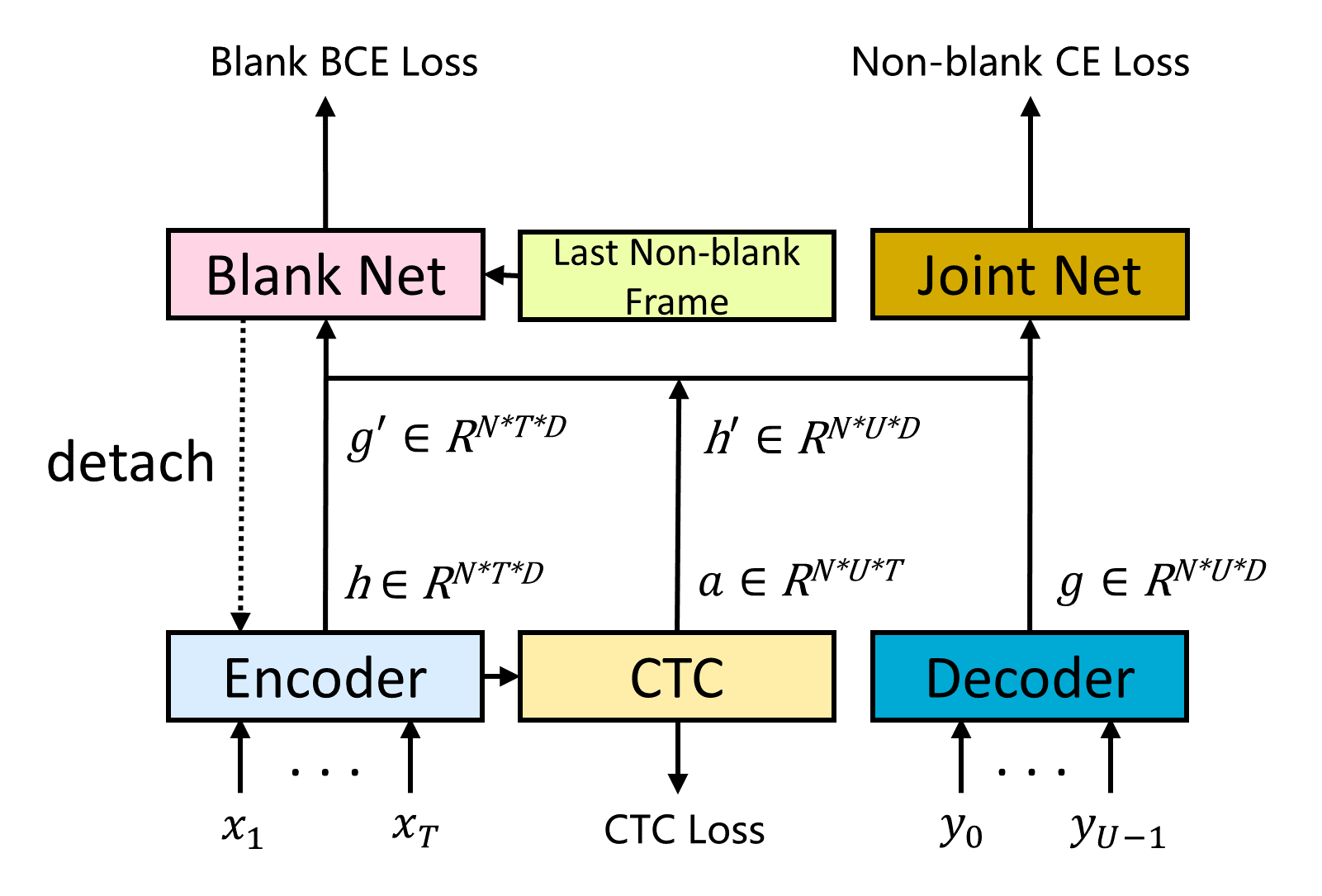}
\caption{The structure of the lightweight transducer. The encoder output \(h\) is transformed to \(h'\) according to CTC forced alignment result \(a\), and then combined with the decoder output \(g\) to input into the joint network. The decoder output \(g\) is also transformed to \(g'\) according to \(a\), and then combined with \(h\) to input into the blank network.}
\label{fig:lightweight transducer}
\end{figure}

The joint network, where each acoustic feature and each language feature are added together, is a major cause of high memory consumption. However, if we know which symbol corresponds to each frame, we only need to consider the symbol history at the corresponding time for each frame, reducing the memory consumption by one order of magnitude. In order to enable end-to-end training of the entire model, we use the CTC forced alignment algorithm to determine which symbol corresponds to each frame. The structure of the lightweight transducer is shown in Fig. \ref{fig:lightweight transducer}.

\subsection{CTC forced alignment}

The CTC modifies label sequence \(y\) to \(y'\), with blanks added to the beginning and the end and between every pair of labels. The CTC forced alignment algorithm finds the most likely path in the CTC posterior probabilities for a given input audio sequence of length \(T\) and a label of length \(|y'|\). This path is computed using the viterbi algorithm. Let the CTC forward variable \(c(t,u)\) be the log-probability outputting \(y'_{0..u}\) after seen \(x_{0..t}\). It can be computed recursively using \eqref{alpha_ctc}
  \begin{align}
    c(t,u) & =
    \begin{cases}
      c'(t,u)y'(t,u), \hspace{2.6em} {if}~y'_u=b \enspace or \enspace y'_u=y'_{u-2} \nonumber\\
      (c'(t,u)+c(t-1,u-2))y'(t,u), \enspace {otherwise}\label{alpha_ctc}
    \end{cases}\\  
 \end{align}
where
\begin{equation}\label{c'} 
  c'(t,u) =c(t-1,u)+ c(t-1,u-1)
\end{equation}
The last frame \(T\) corresponds to the last label \(y'_U\). We can determine which label \(T-1\) frame corresponds to by comparing three variables \(c(T-1,U)\), \(c(T-1,U-1)\) and \(c(T-1,U-2)\). If \(c(T-1,U)\) is the largest, \(T-1\) frame corresponds to \(y'_U\). If \(c(T-1,U-1)\) is the largest, \(T-1\) frame corresponds to \(y'_{U-1}\). Since the last label \(y'_U\) is the blank, \(T-1\) frame won't corresponds to \(y'_{U-2}\) even if \(c(T-1,U-2)\) is the largest. The labels of all frames can be determined recursively. In the CTC forced alignment algorithm, one symbol may be aligned to multiple consecutive frames, while in transducer, one symbol only corresponds to one frame. Therefore, if a symbol is aligned to multiple consecutive frames in results of the CTC forced alignment algorithm, we keep the symbol of the first frame and set the symbol of subsequent frames to blank. There is an open-source implementation of the CTC forced alignment algorithm in torchaudio, but it only supports batch size equal to 1, resulting in very slow use of this implementation during training. Therefore, we implemented a batch version of the CTC forced alignment algorithm. The algorithm is somewhat complex, so we provide the Python code in Fig. \ref{fig:ctcfa} for easy reproduction. The return value \(a\) is a matrix with a shape of \((N, U, T)\). If \(label[b,u]\) is aligned to frame \(t\), then \(a[b,u,t]\) is 1, otherwise it is 0.

%

\begin{figure}[t]
  \centering
  \begin{python}
def CTC_FA(
    logit, # N*T*V, log_probs of the outputs
    label, # N*U, use -1 to represent padding
    blank): # blank index
    N,T,V=logit.shape
    yl=(label>=0).sum(1)*2+1
    label=torch.cat([torch.full(label.shape  \
      ,blank),label],dim=-1).reshape(N,2,-1) \ 
      .permute(0,2,1).reshape(N,-1)
    label=F.pad(label,[0,1],value=blank)
    U=label.shape[1]
    minv=torch.finfo(logit.dtype).min
    bi=torch.arange(N)
    bi2=bi.unsqueeze(-1).repeat(1,U).view(-1)
    dp=torch.full((N,T,U), minv)
    dp[:,0,0]=logit[:,0,blank]
    dp[:,0,1]=logit[bi,0,label[:,1]]
    bp=arange(U).view(1,1,-1).repeat(N,T,1)    
    con=(label==blank)|                     \
      (label==F.pad(label,pad=[2,-2],value=-1))
    for i in range(1,T):
        a=torch.cat([dp[:,i-1,:,None],      \
          F.pad(dp[:,i-1],pad=[1,-1],       \ 
          value=minv)[:,:,None]],dim=-1)
        _,idx1=a.max(-1)
        a=torch.logsumexp(a,-1)
        a2=torch.cat([dp[:,i-1,:,None],     \ 
          F.pad(dp[:,i-1],pad=[1,-1],       \ 
          value=minv)[:,:,None],            \ 
          F.pad(dp[:,i-1],pad=[2,-2],       \ 
          value=minv)[:,:,None]],dim=-1)
        _,idx2=a2.max(-1)
        a2=torch.logsumexp(a2,-1)
        dp[:,i]=logit[bi2,i,label.view(-1)] \ 
          .view(a.shape)+torch.where(con,a,a2)
        bp[:,i]=bp[:,i]-where(con,idx1,idx2)
    a=bp.new_zeros(bp.shape)
    wi=torch.where(dp[bi,-1,yl-1]>          \ 
      dp[bi,-1,yl-2],yl-1,yl-2)
    for i in range(T-1,-1,-1):
        a[bi,i,wi]=1
        wi=bp[bi,i,wi]
    return a[:,:,1::2].permute(0,2,1)
\end{python}
\caption{Python code for the CTC forced alignment algorithm.}
\label{fig:ctcfa}
\end{figure}

\subsection{Decoupling blank probability}

We found that directly using the results of the CTC forced alignment as the label of the lightweight transducer for cross entropy training can result in many deletion and substitution errors. Due to the fact that most of the frames in the CTC forced alignment result correspond to blanks, it can lead to very imbalanced multi classification, resulting in the model predicting too many blanks. To solve this problem, we first decouple blank from multi-classification and use a separate binary classifier \cite{Chen2021FactorizedNT, Zhao2022FastAA} to predict the probability of blank. For non-blank categories, we use a multi-classification classifier. The blank classifier calculates loss for all frames, while the non-blank classifier only calculates loss for non-blank frames. When decoding, the probabilities of blank \(P_b\) and non-blank categories \(P_{nb}\) are combined into a probability distribution \(P\) through interpolation.
\begin{equation}\label{3} 
  P=\{P_b,P_{nb}*(1-P_b)\}
\end{equation}
Besides, we found that it is necessary to truncate the gradient of the blank classifier to the main network, otherwise the effect is still not ideal. The total loss \(L\) is the weighted sum of the CTC loss \(L_{CTC}\), the blank binary classification loss \(L_b\), and the non-blank multi-class loss \(L_{nb}\). Since the label quality of the lightweight transducer depends on the CTC, we do not calculate the lightweight transducer loss when the CTC loss is high.
  \begin{align}\label{5}
    L & =
    \begin{cases}
      \lambda L_{CTC}+(1 - \lambda)L_{nb}+L_b, & \mbox{if}~L_{CTC}<2 \nonumber\\
      L_{CTC}, & \mbox{otherwise}
    \end{cases}\\  
 \end{align}

\subsection{Enhanced blank classifier}

In order to improve accuracy in predicting the blank probability, we analyzed when the model should output a blank: firstly, the current acoustic frame is silent, noise or information is insufficient to determine the symbol. In this case, we can determine whether it is a blank based on the current audio frame only. Secondly, a symbol corresponds to multiple consecutive frames, and the symbol corresponding to the current frame has already been output in the previous frame. At this point, it is necessary to output a blank, otherwise there will be duplicate output. In this case, the first step is to determine whether the symbol corresponding to the current frame is the same as the last emitted symbol. If not, it is not blank. Otherwise, it may be multiple frames corresponding to the same symbol. This needs to be determined by comparing the frame corresponding to the last emitted symbol with the current frame. Therefore, our blank classifier input includes three pieces of information: the current acoustic frame, the current language feature and the acoustic frame corresponding to the last emitted symbol. Compared with the traditional transducer, we includes the last non-blank frame. The experimental results showed that incorporating this information can help the model make more accurate blank judgments, leading to lower CER. Unlike \cite{Xu2022MultiBlankTF}, they focus on predicting big blanks to accelerate inference, while we use richer information to improve the accuracy of blank prediction.

\section{Experiments}

\subsection{Dataset}

In this work, all experiments are carried out on the public Mandarin speech corpus AISHELL-1. The training dataset encompasses approximately 150 hours of speech, consisting of 120,098 utterances contributed by 340 unique speakers. The development set contains around 20 hours of recordings, with a total of 14,326 utterances recorded by 40 speakers. Lastly, the test set consists of about 10 hours of speech (7,176 utterances) recorded by 20 speakers, and the speakers in each of these sets do not overlap.

\subsection{Experimental Setup}

We compared the performance of three models: LAS, transducer, and lightweight transducer. All models have the same encoder and are jointly trained with CTC loss. CTC weight is set to 0.3. The speech features are 80-dimensional log-mel filterbank. We apply SpecAugment \cite{Park2019SpecAugmentAS} with mask parameters (\(F=10, T=50, m_F=m_T=2\)) and speed perturbation with factors of 0.9, 1.0 and 1.1 to augment training data. The encoder is composed of a subsampling module and 12-layer conformer blocks \cite{Gulati2020ConformerCT}. The subsampling module contains two kernel size 3×3 and stride size 2×2 convolution layers with 64 channels. For conformer blocks, the attention dimension is 256 with 4 attention heads, the dimension in the feed-forward network sublayer is 2048. After the fourth layer of conformer, there is a Conv1d with a kernel size and a stride size of 2×2, resulting in 80ms framerate. The decoder is one layer of LSTMP \cite{Sak2014LongSM} with a hidden size of 1024 and a projection size of 512. The cross attention of the LAS is computed over 128-dimensional location sensitive attention \cite{Chorowski2015AttentionBasedMF}. The blank network is composed of two layers of linear, with dimensions of 256 and 1, and activation functions of tanh and sigmoid, respectively. We apply the Adam optimizer \cite{Kingma2014AdamAM} with a gradient clipping value of 5.0 and warmup \cite{Vaswani2017AttentionIA} learning rate scheduler with 25,000 warmup steps and peak learning rate 0.0015. We train the models for 180 epochs and average the last 30 checkpoints to obtain the final model. Our model was trained on one A100 80GB GPU with an actual batch size 256. Specifically, the batch size of LAS and lightweight transducer is 128, and the gradient accumulation is 2. The batch size and gradient accumulation of transducer are both 16 because setting the batch size to 32 will cause OOM problems. We use torchaudio \cite{yang2021torchaudio} to calculate the transducer loss.

\section{Results}

\subsection{Overall results}

Table 1 compares training hours and CERs of different models. The training time of the lightweight transducer is similar to that of LAS and significantly shorter than that of transducer. Since lightweight transducer significantly reduces the training memory requirement, the batch size can be set to 128, which is eight times that of the transducer. And lightweight transducer can achieve similar results to LAS, superior to transducer. Although the accuracy of the lightweight transducer is slightly lower than that of LAS, LAS cannot decode audio in real-time due to its reliance on global context, while the lightweight transducer can.

\subsection{Ablation studies}

Table 2 verifies the effectiveness of each component of the lightweight transducer. If the last non-blank frame is not input into the blank classifier, the effect decreases to the same level as the transducer, indicating that adding this information can make the probability of blank classification more accurate. And if the gradient of the blank classifier to the main network is not truncated, the effect will significantly decrease. This indicates that the gradient of the blank classifier has a negative effect to the main network. Although decoupling the blank probability does not improve the effect, it is necessary for other components.

\subsection{Comparison on robustness}
Although Table 1 shows that the LAS model has the highest accuracy rate, the LAS model has a well-known problem of poor robustness, especially when decoding long audio clips, it is easy to have repeated decoding or word dropping problems. Therefore, we conducted an experiment on long utterances recognition. We spliced adjacent audio clips in the test set for decoding. The result is shown in Table 3. Compared with Table 1, the WER of all models has significantly increased, with substitution errors rising notably. This is primarily because the concatenated audio segments have no inherent relationship with each other, and the models are affected by irrelevant contexts. Besides, the LAS model is more susceptible to the mismatched audio lengths between training and testing, even though we used location sensitive attention. The results show that lightweight transducer has the same robustness as transducer, significantly better than LAS, and has the lowest WER on long utterances.

\begin{table}[th]
\centering
\caption{Results of different models}
\begin{tabular}{lccc}
   \toprule
   Model & training hours & Dev & Test \\
   \midrule
   LAS & 18.1 & 4.26 & 4.67 \\
   Transducer & 64.2 & 4.59 & 5.07 \\
   Lightweight Transducer & 18.6 & 4.45 & 4.76 \\
   \bottomrule
\end{tabular}
\end{table}

\begin{table}[th]
\centering
\caption{Results of ablation studies}
\begin{tabular}{lcccc}
   \toprule
   Model & Ins & Del & Sub & Test \\
   \midrule
   LAS & 0.11 & 0.14 & 4.43 & 4.67 \\
   Transducer & 0.13 & 0.24 & 4.71 & 5.07 \\
   \midrule
   Lightweight Transducer & 0.08 & 0.20 & 4.49 & 4.76 \\
   \hspace{1em} -Enhanced Blank Classifier & 0.10 & 0.27 & 4.71 & 5.07 \\
   \hspace{2em} -Truncated Gradient & 0.13 & 0.35 & 7.29 & 7.76 \\
   \hspace{3em} -Decoupling Blank & 0.17 & 0.36 & 6.85 & 7.37 \\
   \bottomrule
\end{tabular}
\end{table}

\begin{table}[th]
\centering
\caption{Results on long utterances. The number in the cat column represents the number of spliced audio.}
\begin{tabular}{lccccc}
   \toprule
   Model & Cat & Ins & Del & Sub & WER \\
   \midrule
   \multirow{3}{*} {LAS} & 2 & 0.12 & 0.74 & 8.06 & 8.91 \\
   & 4 & 0.76 & 14.90 & 11.30 & 26.95 \\
   & 8 & 1.44 & 39.85 & 16.63 & 57.92 \\
   \midrule
   \multirow{3}{*} {Transducer} & 2 & 0.15 & 0.59 & 8.80 & 9.55 \\
   & 4 & 0.16 & 0.88 & 11.36 & 12.40 \\
   & 8 & 0.20 & 1.09 & 13.14 & 14.42 \\
   \midrule
   \multirow{3}{*} {\shortstack{Lightweight \\ Transducer}} & 2 & 0.11 & 0.37 & 8.38 & 8.86 \\
   & 4 & 0.11 & 0.53 & 11.44 & 12.08 \\
   & 8 & 0.10 & 0.63 & 13.30 & 14.03 \\
   \bottomrule
\end{tabular}
\end{table}

\section{Conclusion}

In this work, we introduced lightweight transducer. By using CTC forced alignment results as labels and making a series of improvements, we successfully trained the lightweight transducer model based on frame-level cross entropy loss. The experimental results showed that the lightweight transducer exhibits training speeds and accuracy comparable to LAS, significantly outperforms the traditional transducer, and demonstrates superior robustness over LAS.

\section{Acknowledgements}
This work was supported by the National Key R\&D Program of China (2022YFB4500600).

\bibliographystyle{IEEEtran}
\bibliography{mybib}

\end{document}